\documentclass[10pt,twocolumn,letterpaper]{article}

\usepackage[pagenumbers]{cvpr} 

\usepackage{graphicx}
\usepackage{amsmath}
\usepackage{amssymb}
\usepackage{dsfont}
\usepackage{booktabs}
\usepackage{wrapfig}
\usepackage{bm}
\usepackage{algorithm}
\usepackage{algpseudocode}
\usepackage{enumitem}

\usepackage[pagebackref,breaklinks,colorlinks]{hyperref}

\usepackage[capitalize]{cleveref}
\crefname{section}{Sec.}{Secs.}
\Crefname{section}{Section}{Sections}
\Crefname{table}{Table}{Tables}
\crefname{table}{Tab.}{Tabs.}

\def\cvprPaperID{2886} 
\def\confName{CVPR}
\def\confYear{2023}

\begin{document}

\title{\MethodName: Unpaired Structure-guided Masked Image Generation}
\newcommand{\MethodName}{MaskSketch}
\newcommand{\DatasetName}{OpenSketches}

\author{Dina Bashkirova\thanks{Boston University} \thanks{This work was done during an internship at Google.}\\

\and
José Lezama\thanks{Google Research}

\and
Kihyuk Sohn\footnotemark[3]
\and
Kate Saenko\footnotemark[1]
\and
Irfan Essa\thanks{Georgia Institute of Technology} \footnotemark[3]}
\maketitle

\begin{abstract}
   
   Recent conditional image generation methods produce images of
   remarkable diversity, fidelity and realism. However, the majority
   of these methods allow conditioning only on labels or text prompts,
   which limits their level of control over the generation result. In
   this paper, we introduce \MethodName, an image generation
   method that allows spatial conditioning of the generation result
   using a guiding sketch as an extra conditioning signal during sampling. {\MethodName}
   utilizes a pre-trained masked generative transformer, requiring no model
   training or paired supervision, and works with input sketches of
   different levels of abstraction. 
   We show that intermediate self-attention maps of a masked generative transformer encode important structural information of the input image, such as scene layout and object shape, and we propose a novel sampling method based on this observation to enable structure-guided generation.
   Our results show that
   {\MethodName} achieves high image realism and fidelity to the guiding
   structure. Evaluated on standard benchmark datasets, {\MethodName}
   outperforms state-of-the-art methods for sketch-to-image
   translation, as well as unpaired image-to-image translation
   approaches.
   
\end{abstract}

\begin{figure}[t]
    \centering
    \includegraphics[width=\linewidth]{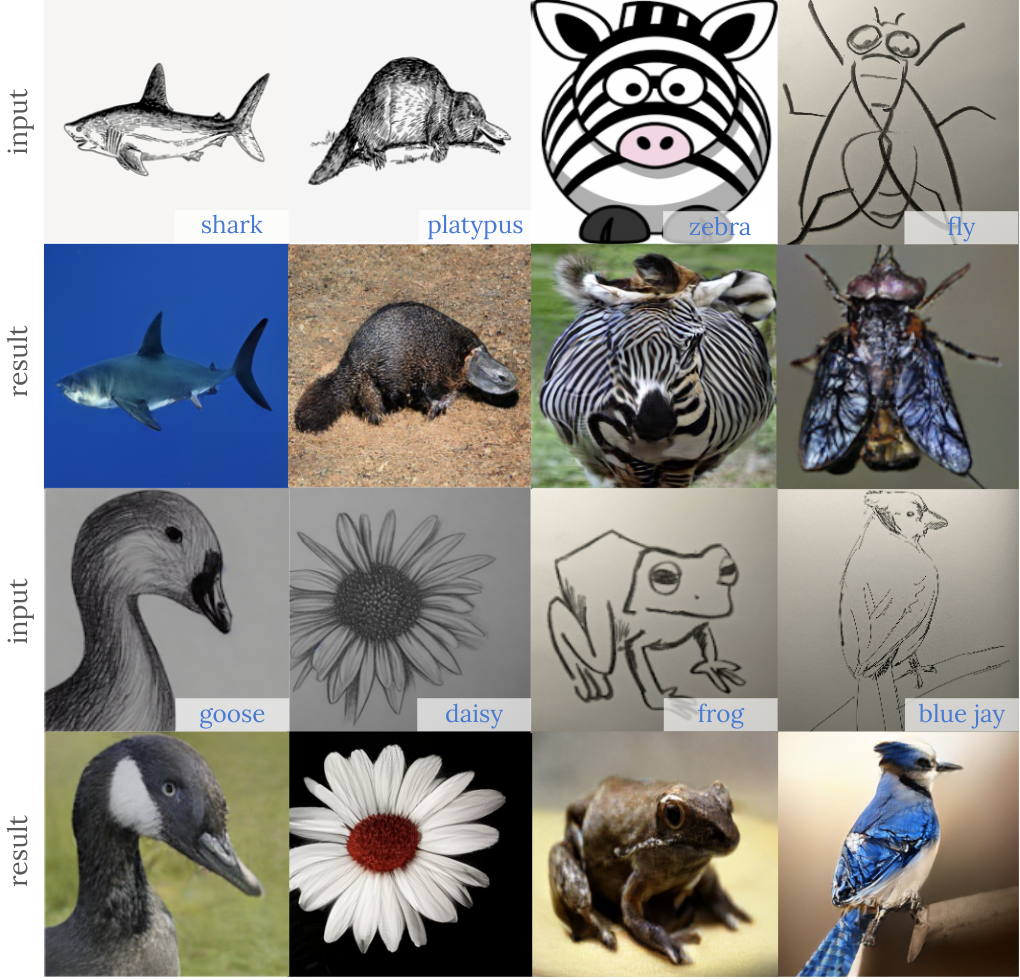}
    \caption{Given an input sketch and its class label, {\MethodName} samples realistic images that follow the given structure. {\MethodName} works on sketches of various degrees of abstraction by leveraging a pre-trained masked image generator \cite{chang2022maskgit}, while not requiring model finetuning or pairwise supervision.  \vspace{-5pt}}
    \label{fig:results_vis}
\end{figure}

\section{Introduction}
\label{sec:intro}
Image generation methods recently achieved remarkable success, allowing diverse and photorealistic image synthesis~\cite{saharia2022photorealistic,chang2022maskgit,dhariwal2021diffusion,rombach2022high}. The majority of state-of-the-art generative models allow conditioning with class labels~\cite{brock2018large,esser2020taming,chang2022maskgit,dhariwal2021diffusion} or text prompts~\cite{ramesh2021zero, ramesh2022hierarchical, saharia2022photorealistic, rombach2022high}, however, applications require a more fine-grained control over the spatial composition of the generation result. 
 While methods that use conditioning with segmentation maps~\cite{gafni2022make} or strokes~\cite{meng2022sdedit} achieve some spatial control over the generated image, sketching allows a more fine-grained specification of the target spatial layout, which makes it desirable for many creative applications.


In this paper, we propose \MethodName, a method for conditional image synthesis that uses sketch guidance to define the desired structure, and a pre-trained state-of-the-art \emph{masked generative transformer}, MaskGIT~\cite{chang2022maskgit}, to leverage a strong generative prior. We demonstrate the capability of {\MethodName} to generate realistic images of a given structure for sketch-to-photo image translation. Sketch-to-photo
~\cite{ham2022cogs,chen2018sketchygan,lu2018image} is one of the most challenging applications of structure-conditional generation due to the large domain gap between sketches and natural images. 
{\MethodName} achieves a balance between realism and fidelity to the given structure. Our experiments show that {\MethodName} outperforms state-of-the-art sketch-to-photo~\cite{ham2022cogs} and general unpaired image translation methods~\cite{huang2018munit,park2020contrastive,chen2022eccv}, according to  standard metrics for image generation models~\cite{heusel2017gans} and user preference studies.

\begin{figure}[t]
  \begin{center}
    \includegraphics[width=\linewidth]{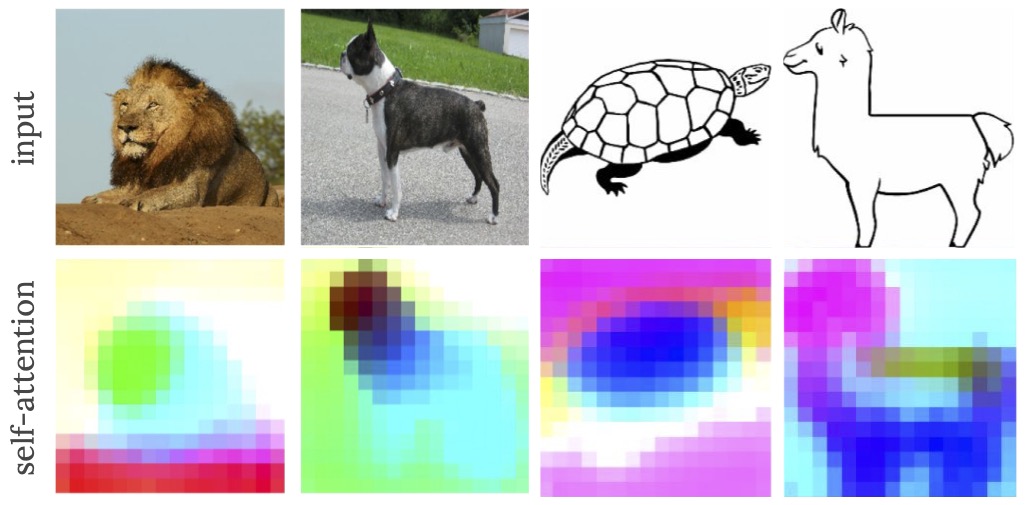}\vspace{-10pt}
  \end{center}
  \caption{ Self-attention maps (PCA) of the intermediate layers of a pre-trained masked generative  transformer \cite{chang2022maskgit} encode information about the spatial layout of the input. Notably, they are robust to the domain shift between natural images (\textbf{left}) and sketches (\textbf{right}).  \vspace{-10pt}}
  \label{fig:attn_fig}
\end{figure}

 
In {\MethodName}, we formulate a structure similarity constraint based on the observation that the intermediate self-attention maps of a generative transformer~\cite{chang2022maskgit} encode rich structural information  (see \cref{fig:attn_fig}). We  use this structure similarity constraint
to guide the generated image towards the desired spatial layout~\cite{Splice2022,hertz2022prompt}. Our study shows that the proposed attention-based structure similarity objective is robust to the domain shift occurring in sketch-to-photo translation.
The proposed structure-based sampling method leverages a pre-trained image generator, and does not require model finetuning or sketch-photo paired data. Moreover, it is significantly faster than other methods that exploit self-attention maps for guided image generation~\cite{Splice2022}. 
 %
%
  Figure~\ref{fig:results_vis} shows the translation results produced by our method on sketches of various levels of abstraction.
  
The limitations of existing sketch-to-photo translation
methods~\cite{chen2018sketchygan, ham2022cogs, lu2018image} come from
having to learn both an implicit natural domain prior and the mapping
that aligns sketches to natural images, for which the domain gap is
severe. {\MethodName}, on the other hand, uses the strong generative
prior of a pre-trained generative transformer, which allows highly
realistic generation. In addition, {\MethodName} uses the
domain-invariant self-attention maps for structure conditioning,
allowing its use on sketches of a wide range of abstraction levels.
  
Our contributions can be summarized as follows:
\begin{itemize}[noitemsep,nolistsep]
    \item We show that the self-attention maps of a masked generative transformer encode important structural information and are robust to the domain shift between images and sketches.
    \item We propose a sampling method based on  self-attention similarity, balancing the structural guidance of an input sketch and the natural image prior. 
    \item We demonstrate that the proposed sampling approach, \MethodName, outperforms state-of-the-art methods in unpaired sketch-to-photo translation.
    \item  To the best of our knowledge, {\MethodName} is the first method for sketch-to-photo translation in the existing literature that produces photorealistic results requiring only class label supervision.
\end{itemize}

\section{Related Work}
\label{sec:related}

While there is a vast volume of literature on image generative models thanks to recent progress ranging from generative adversarial networks~\cite{goodfellow2020generative,brock2018large,karras2019style}  generative transformers~\cite{esser2020taming,chang2022maskgit,yu2022scaling} and diffusion models~\cite{dhariwal2021diffusion, ramesh2022hierarchical,nichol2021glide,saharia2022photorealistic}, in this section, we focus on reviewing image-conditioned image generation, also known as image translation.

\vspace{-10pt}\paragraph{Supervised image conditional generation}
Sketch-to-photo image translation is a special case of image-conditional image generation.
Early conditional image generation methods were based on generative adversarial networks. For example, pix2pix~\cite{isola2017image} conditioned the generation result by minimizing the distance between the ground truth and the generated image; SPADE~\cite{park2019SPADE} and OASIS~\cite{sushko2020you} used spatially-adaptive instance normalization to condition the resulting image on a segmentation map; CoCosNet~\cite{zhang2020cross} and CoCosNet V2~\cite{Zhou_2021_CVPR} warped the real reference image using a correlation matrix between the image and the given segmentation map. Similarly to {\MethodName}, Make-a-Scene and NUWA~\cite{gafni2022make,wu2021n} use a VQ-based transformer architecture and is designed to condition generation on semantic segmentation and text prompts. While these methods allow spatial conditioning, they are inapplicable for sketch-to-photo due to the lack of ground truth paired data, domain gap between sketches and segmentation maps and lack of efficient methods that extract semantic segmentation from sketches. 
 
\begin{figure*}
    \centering
    \vspace{-15pt}
    \includegraphics[width=\textwidth]{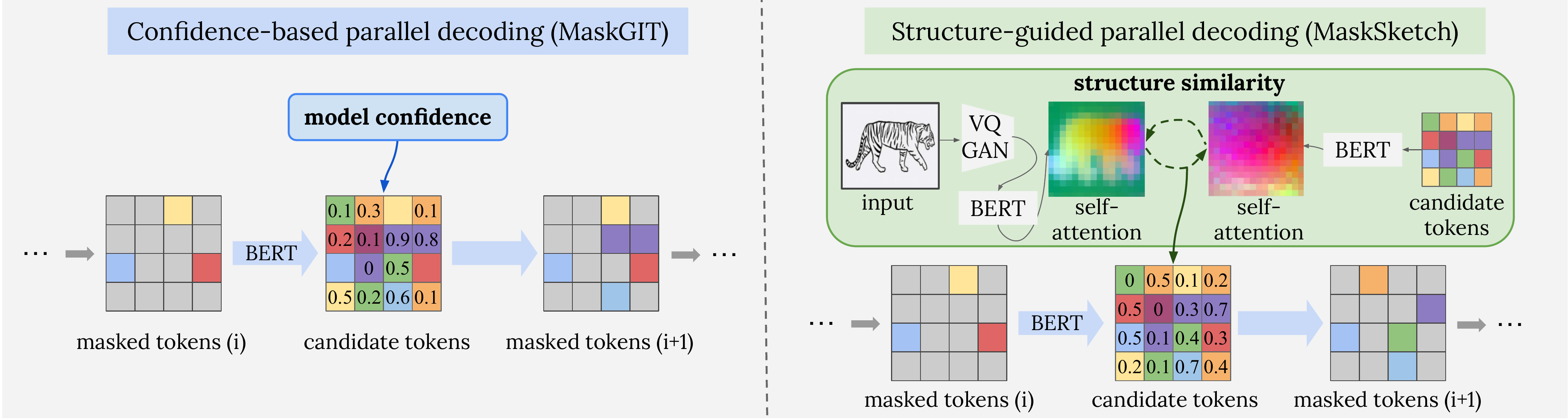}
    \caption{  \textbf{Left:}  confidence-based token rejection (masking) in MaskGIT. \textbf{Right:} structure-based token rejection. Confidence-based rejection masks out the least `likely' tokens, while structure-based rejection masks the tokens with the highest structural distance (Eq. \ref{eq:structure_dist}) w.r.t. the input sketch.}  \vspace{-5pt}
    \label{fig:method_schematic}
\end{figure*}

\vspace{-10pt}\paragraph{Unsupervised image-conditional generation}
In unsupervised image-conditioned translation, the ground truth input and translation pairs are not available for training. For example, CycleGAN~\cite{zhu2017unpaired} used a cycle reconstruction loss to ensure a semantically consistent translation, UNIT~\cite{liu2017unsupervised}, MUNIT~\cite{huang2018munit}, and StarGANv2~\cite{choi2020starganv2}  disentangled domain-specific and shared information between the source and target image domains by mapping them to a shared latent embedding space. PSP~\cite{richardson2021encoding} used StyleGAN~\cite{saharia2022photorealistic} inversion along with style mixing for segmentation map- and edge-guided image translation. SDEdit~\cite{meng2021sdedit} uses a diffusion model to translate the input strokes or segmentation maps to natural images.

The closest work to ours in this line may be Splice-VIT~\cite{Splice2022}, which uses self-attention key self-similarity extracted from the discriminative ViT (Dino \cite{caron2021emerging}) to represent the structure of an input image. 
As pointed in~\cite{Splice2022},
Splice-VIT works only in case when both the input and expected output images come from the same domain, which makes it inapplicable for sketch-to-photo translation. VQ-I2I~\cite{chen2022eccv} is another work on  unsupervised image translation that leverages the generative power of a VQ-GAN~\cite{esser2020taming}-based generative transformer. Unlike \MethodName, VQ-I2I uses the embedding reconstruction loss for controllable generation. 

Recently, \cite{kawar2022imagic,gal2022image, ruiz2022dreambooth} also demonstrated how to leverage pre-trained image generators for image synthesis based on novel conditioning inputs. These methods allow to replicate a given object or subject in the generated image. However, they do not allow  to finely specify the spatial layout of the generated image as  {\MethodName} and typically require some degree of fine-tuning. Prompt-to-prompt tuning~\cite{hertz2022prompt} uses the attention features to perform spatially aligned prompt-conditional generation. 
General image-conditional methods show remarkable results when the source and target domains are visually similar, e.g., translating horses to zebras, performing artistic style transfer, etc, however, they tend to struggle on the more challenging sketch-to-photo translation task. {\MethodName} shows a promising alternative for transferring the spatial composition from sketches of various degrees of abstraction, since it requires no paired data or model training, thanks to leveraging a powerful pre-trained generator.


\vspace{-10pt}\paragraph{Sketch-to-photo translation} 
The sketch-to-photo application received attention in recent years thanks to the advancement in the field of image generation.
For example, SketchyGAN~\cite{chen2018sketchygan} proposes an a GAN-based approach based on edge-preserving image augmentations, ContextualGAN~\cite{lu2018image} leverages conditional GAN along with joint image-sketch representation, iSketchNFill~\cite{ghosh2019interactive} uses a gating mechanism to condition output images on the class label and an MUNIT-based generator to synthesize images with diverse appearance. Photosketcher~\cite{eitz2011photosketcher} uses sketch-based image retrieval to compose a real image. PITI~\cite{wang2022pretraining} is pretrained with ground truth edge maps and semantic segmentation maps to learn a domain invariant semantic representation.  The state-of-the-art supervised method CoGS~\cite{ham2022cogs} minimizes the distances between the structure embeddings of the input sketch and the corresponding ground truth real image in the vector-quantized space of a VQ-GAN~\cite{esser2020taming}.
In contrast, {\MethodName}  does not rely on paired data for training, which allows it to use sketches of different abstraction levels, as well as real photos.


\section{Method}
\label{sec:method}
In this section, we describe the main components in {\MethodName} that introduce sketch-guided spatial control to a conditional masked image generator. 
We first review masked image generation in Section~\ref{sec:background}. Then we introduce the two main components of \MethodName, a structure similarity distance in Section~\ref{sec:struct_score}, and structure-guided parallel sampling in Section~\ref{sec:decoding}.
Finally, we discuss how to balance the trade-off between structure fidelity and generation realism in Section~\ref{sec:fidelity_realism}. 

\subsection{Background: Masked Image Generation}\label{sec:background}
Masked image generation is a state-of-the-art approach for efficient generation~\cite{chang2022maskgit,lezama2022token,gu2021vector}, combining the strengths of masked token modeling~\cite{devlin2018bert} and non-autoregressive sampling~\cite{ghazvininejad2019mask}. It encodes images as discrete sequences of visual tokens using a VQ-GAN encoder~\cite{esser2020taming}, and then trains a bi-directional transformer (BERT~\cite{devlin2018bert}) to model natural image distribution in the discrete token sequence space. Generation is performed iteratively, where significant gains in efficiency are obtained by using parallel sampling instead of auto-regressive sampling. 
MaskGIT~\cite{chang2022maskgit} starts from a blank canvas where all visual tokens are masked. At each sampling iteration, all the missing tokens are sampled in parallel, and a rejection criteria is used, where the tokens with low model likelihood are masked and will be re-predicted in the next refinement iteration. See Figure~\ref{fig:method_schematic} (left) for an illustration of a single MaskGIT decoding step. 
{\MethodName} extends the parallel sampling of MaskGIT to sample images that follow the structure determined by an input image (\cref{fig:method_schematic}, right), as described in the following sections. 

\subsection{Structure Similarity via Attention Maps}
\label{sec:struct_score}
We consider two images to be structurally similar when their self-similarity maps are close to each other. {\MethodName} leverages the  self-similarity encoded in the self-attention maps of a masked generative transformer (Section~\ref{sec:background}) to define structural distance. One key observation in our work is that a class-conditional MaskGIT trained on ImageNet shows a high degree of domain invariance in its attention maps and is able to capture the self-similarity in out-of-distribution domains such as sketches (\cref{fig:attn_fig}). 

Formally, we define a structural distance based on a comparison of self-attention maps. 
Let $\mathds{Z}$ be the indices representing the VQ-GAN~\cite{esser2020taming} dictionary of vector-quantized image tokens.
Let $\bm{x}\in \mathds{Z}^N$ be sequence of $N$ discrete tokens obtained using a VQ-GAN encoding an input image in the vector-quantized space.
Given an input image $\bm{x}$ and a generated image $\bm{y}$, let $A^\ell(\bm{x})\in {\left[0,1\right]}^{N\times N}$ be the transformer self-attention map at layer $\ell$.  Each row in $A^\ell(\bm{x})$ represents the attention weights of each token with respect to all tokens, normalized with a softmax function. We define the structural distance between the $i
^\text{th}$ tokens of images $\bm{x}$ and $\bm{y}$ across layers $\mathcal{L}$ as:
\begin{equation}
    d_S^i(\bm{x}, \bm{y}) = \sum_{\ell \in \mathcal{L}}   d_{J}\left(A_i^\ell(\bm{x}), A_i^\ell(\bm{y}) \right),
    \label{eq:structure_dist}
\end{equation}
where $d_{J}$ is the Jeffrey's divergence:
\begin{equation}
    d_{J}(\bm{u}, \bm{v}) = \frac{KL\left(\bm{u} \Vert \bm{v} \right) + KL\left(\bm{v} \Vert \bm{u} \right)}{2}.
\end{equation}
Intuitively, the image regions represented by the $i^{\text{th}}$ tokens of  $\bm{x}$ and $\bm{y}$ are structurally similar if their distributions of attention self-similarities are close to each other.
\begin{figure}[t]
    \centering
    \includegraphics[width=1.\linewidth]{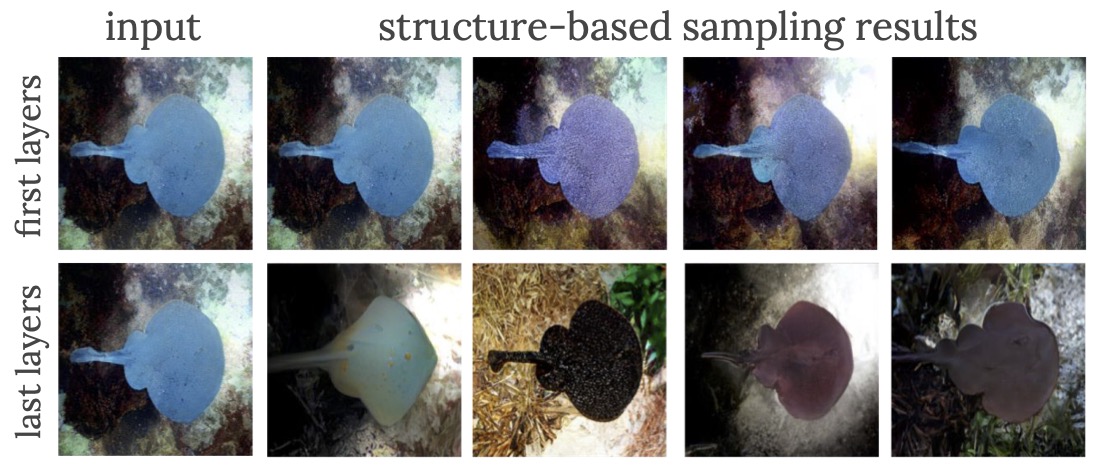}
    \caption{Structure-guided sampling  using attention maps from the first three layers of  a masked generative transformer  results in a nearly perfect reconstruction of the input, whereas using  the final layers $(16, 18, 20)$, out of $24$, yields realistic images with a similar structure but highly diverse  appearance. 
    (\textit{Best viewed in color}.)  \vspace{-5pt}}
    \label{fig:sampling_example_towels}
\end{figure}
\subsection{Structure-guided Parallel Decoding}
\label{sec:decoding}
{\MethodName} adapts the parallel sampling of MaskGIT to take into account the structural similarity between the output and the reference input sketch. More precisely, the token rejection criteria in each decoding iteration is modified to also reject the sampled tokens that have low self-similarity  score  ~\eqref{eq:structure_dist}. The proposed structure-guided decoding strategy can also be seen as a greedy optimization technique that balances minimizing the structural distance and following the model's image prior. 

While MaskGIT sampling rejects token candidates with the lowest likelihood by masking them at the end of each decoding iteration, {\MethodName} creates an additional mask that rejects tokens based on the structural similarity to the input sketch (see~\cref{fig:method_schematic}). At the end of one decoding iteration, we compute the logical OR between the confidence-based and the structure-based masks to optimize both realism and structure similarity
~(\cref{sec:fidelity_realism}).  The pseudocode of our algorithm is described in Algorithm~\ref{algo:sampling}. It relies on the function {\tt sample\_mask}, which takes as input a vector of structure similarity scores  $\bm{s}^s$ and the number of masked elements $k$, and samples a mask by Gumbel top-$k$ using  $\bm{s}^s$ to  mask the tokens with the highest structure distance.

\begin{figure}   
    \begin{tabular}{c}
%
   \includegraphics[width=1.\linewidth]{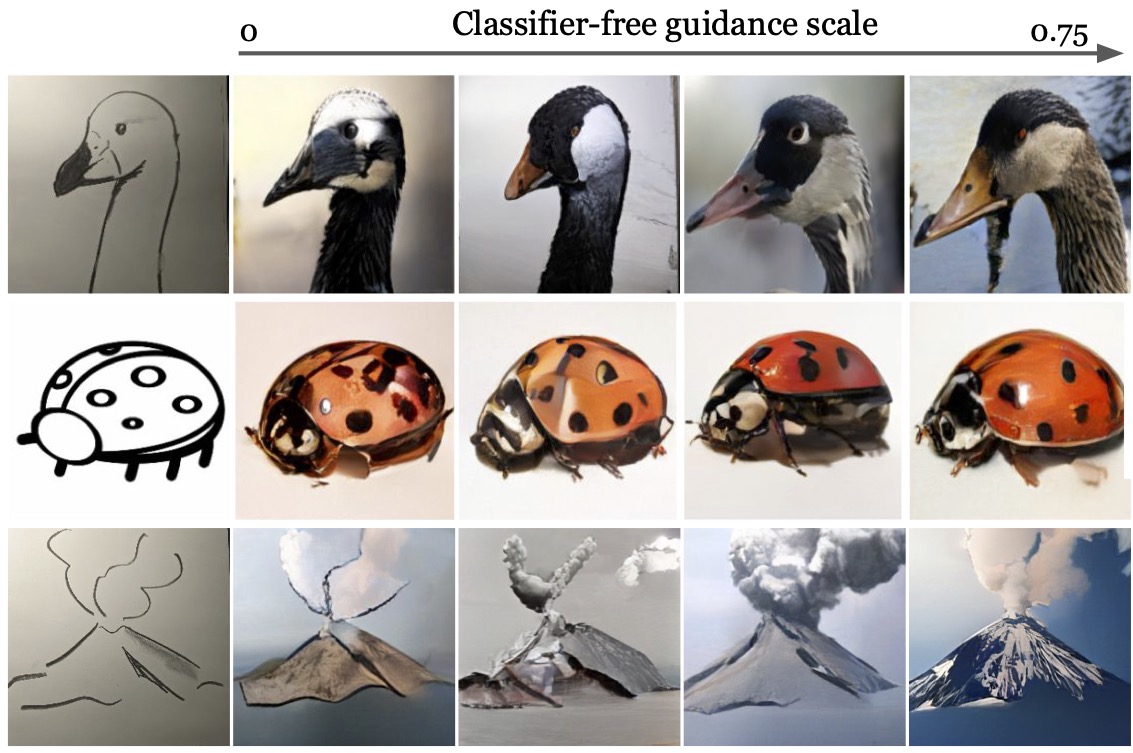}\\
   %
   \includegraphics[width=\linewidth]{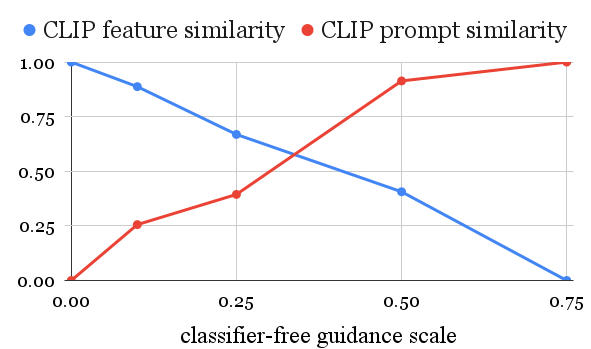}
   \end{tabular}
\hfill
\vspace{-0pt}
\caption{\textbf{Top:} Traversal of the realism-fidelity trade-off by varying the classifier-free guidance scale. \textbf{Bottom:}  Increasing the guidance scale leads to higher realism and a high CLIP similarity to the prompt ``photo of a \textit{c}'', where \textit{c} is the class name, at the cost of lower structure fidelity and lower CLIP feature similarity. \vspace{-15pt}}
 \label{fig:realism_fidelity}
\end{figure}

The choice of layers $\mathcal{L}$ selected for computing the structure similarity significantly impacts the generation results. Our experiments show that sampling with the attention maps extracted from the first layers  results in nearly identical reconstruction of the given input image, whereas minimizing the structure distance based on the last layers results in images of diverse appearance that are spatially aligned with the input image at a high level, as shown in~\cref{fig:sampling_example_towels}. 

\begin{algorithm}[b]
    \caption{{\MethodName} sampling}\label{algo:sampling}
     \noindent\textbf{Input:} Pre-trained BERT generator $G$, structure and confidence masking schedule function  $\gamma(t)$, structure-based sampling ratio $\lambda_{s}$, input sketch $\bm{x}$,  layer(s) $\ell$. \\
     \noindent\textbf{Output:} Generated image encoding $\bm{y}_0$  
    \begin{algorithmic}[1]
    \State {$A^\ell(\bm{x})\gets \text{\tt attn\_map}(G, \bm{x}, \ell)$}
    \State {Initialize $\bm{y}_T$  }
            \For{$t=T-1\ldots 0$}
        \State { $\bar{\bm{y}}_t,~ \bm{s}^{c}_t = G(\bm{y}_{t+1}) $  }
        
        \State {
        $A^\ell(\bar{\bm{y}}_t) = \text{\tt attn\_map}(G, \bar{\bm{y}}_t, \ell)$}
        \State{
        $\bm{s}^{s}_t \gets \left\{ d_J\left(A_i^\ell(\bar{\bm{y}}_t),~ A_i^\ell(\bm{x}) \right) \right\}_{i=1\ldots N} $ 
        }
        \State{
        $\bm{m}^{s}_t = \text{\tt sample\_mask}(\bm{s}^{s}_t, \lfloor\lambda_s\gamma(t) \cdot N \rfloor)$
        }
        \State{
        $\bm{m}^{c}_t = \text{\tt sample\_mask}(\bm{s}^{c}_t, \lfloor(1 - \lambda_s)\gamma(t)\cdot N\rfloor)$
        }
        \State{
        $\bm{m}_t = \bm{m}_t^{s} ~ \lor ~ \bm{m}_t^{c}$
        }
        \State{
        $\bm{y}_t = \bar{\bm{y}}_t \odot \bm{m}_t$
        }
    \EndFor            
    \end{algorithmic}
\end{algorithm}
\subsection{Structure fidelity vs realism trade-off}
\label{sec:fidelity_realism}
One of the biggest challenges in sketch-to-real translation is the immense domain gap between the source and target domains. The input sketches and natural images differ significantly not only in appearance, but also in the distribution of shapes and spatial composition. Due to the domain gap, optimization based solely on the structure distance often results in structurally similar but unrealistic images. To overcome this issue, we propose a combined masking approach that optimizes both structure fidelity and realism. 

To navigate this trade-off, we use a parameter $\lambda_s \in [0, 1]$ to determine the proportion of tokens masked according to the structure similarity scores and those masked according to the model confidence or likelihood scores. Given an overall masking rate schedule function $\gamma(t)$ at step $t$, the structure-based mask rate is computed as $\lambda_s\gamma(t)$, whereas the confidence-based mask rate is $(1 - \lambda_s)\gamma(t)$.
Two independent masks, $\bm{m}_t^{s}$ and $\bm{m}_t^{c}$, are computed for the structure-based and confidence-based scores, respectively. The final mask at iteration $t$ is then computed as the logical OR between $\bm{m}_t^{s}$ and $\bm{m}_t^{c}$.

\vspace{-10pt}\paragraph{Classifier-free Guidance} To further improve the level of realism in the translation result, we use classifier-free guidance~\cite{ho2021classifier, gafni2022make, nichol2021glide} when computing the model likelihood scores. Specifically, for a given sequence of sampled tokens $\bm{\bar{y}}$ and input image $\bm{x}$, we use the pre-trained generator $G$ to compute the per-token logits $\log p(\bm{\bar{y}}(i) |\bm{x}, c)$ conditioned on the correct class $c$ and logits conditioned on a random class $r$: $\log p(\bm{\bar{y}}(i) | \bm{x}, r)$, and calibrate the final confidence-based scores as follows:
\begin{align}
\bm{s}^c(i) =&  \log p(\bm{\bar{y}}(i)| \bm{x}, c) \nonumber\\&- \beta\left(\log p(\bm{\bar{y}}(i) |\bm{x}, c) - \log p(\bm{\bar{y}}(i) | \bm{x},  r)\right)
\end{align}
where $\beta$ is the classifier-free guidance scale.
Figure~\ref{fig:realism_fidelity} shows  how varying $\beta$  affects the fidelity-realism trade-off.
\vspace{-10pt}\paragraph{Global CLIP-based rejection sampling}
Minimization of the structure similarity distance in the space of visual tokens is a discrete optimization problem that cannot be efficiently tackled via continuous optimization methods such as gradient descent. Moreover, since MaskGIT was trained to minimize a different objective, such a greedy optimization process requires more iterations than regular sampling. To increase the stability of the proposed method, we improve the overall fidelity by producing multiple translation samples for a given sketch with different random seeds and guidance scales $\beta$, and selecting the image that yields the highest structure fidelity and realism according to a CLIP-based score. 
Inspired by the recent success in  photo-to-sketch mapping with CLIP~\cite{radford2021learning} domain-invariant representations~\cite{vinker2022clipasso}, we use the $L_1$ distance between features of a CLIP encoder $\operatorname{CLIP}_{s}(\bm{x}, \bm{y})$ of the input image $\bm{x}$ of class $c$ and generation result $\bm{y}$ to estimate the structure similarity. We also use the CLIP similarity score  $\operatorname{CLIP}_{r}(c, \bm{y})$  between the translated image and the corresponding prompt $\operatorname{prompt}(c) =$ `photo of a \textit{c}' to assess the realism for each generated example, more details can be found in ~\cref{sec:sup_clip_metrics}. We normalize the scores across $R$ trials and keep the result with the highest overall quality score: 
\begin{equation} 
\label{eq:clip}
    \bm{y}_{final} = \underset{{\bm{y}} \in \{\bm{y}_1 \ldots \bm{y}_R \}}{\operatorname{argmax}} (1 - \operatorname{CLIP}_{s}(\bm{x}, {\bm{y}}))^2  \operatorname{CLIP}_{r}(c, {\bm{y}})
\end{equation}

~\cref{tab:clip_ablation} in the Appendix shows how the proposed CLIP-based selection approach improves the overall generation result as the number of sampling trials increases.
\section{Experiments}
\label{sec:exp}

\paragraph{Experimental Setup}
In all experiments, we used a class-conditional MaskGIT model pretrained on the ImageNet 2012~\cite{deng2009imagenet} dataset with the output resolution $256{\times}256$. We used layers $1, 3, 16, 20, 21 \text{ and } 22$ to formulate the structure preservation objective. We validated this choice on $100$ random sketches considering structure preservation and realism. In our experiments, for each input sketch, we sample the images four times ($R=4$) with different classifier-free guidance scales (i.e. $\beta \in \{0.0,0.05,0.1,0.25\}$), and select the one that maximizes the CLIP-based objective
in~\cref{eq:clip}. 
We use a linear decay mask rate schedule in all experiments, starting from $\gamma(T)=0.95$ and stopping sampling at the mask rate $\gamma(0)=0.25$, which results in higher realism and reduces artifacts associated with structure-based sampling. To further increase realism, we postprocess the generated samples with Token-Critic refinement~\cite{lezama2022token}, which adds $32$ sampling iterations. See~\cref{sec:token_critic} for more details. We generate each image using $T=500$ sampling iterations, and the overall sampling time for a batch of $8$ images is on average $750$ seconds on a single TPUv4, including four trials and the CLIP-based evaluation.

\paragraph{Baselines}
We consider well-established \emph{unpaired} image-to-image translation methods as baselines. Specifically, we used CUT~\cite{park2020contrastive},  MUNIT~\cite{huang2018munit} and VQI2I~\cite{chen2022eccv} in our comparisons. 
We note that for sketch-to-photo translation, methods that use ground truth attribute information to translate from one attribute to another, e.g. StarGAN~\cite{choi2018stargan, choi2020starganv2}, fail to minimize the gap between sketch and real domains~\cite{ham2022cogs}.   
CUT~\cite{park2020contrastive} uses a contrastive objective to ensure structural similarity between the corresponding patches of the input image and the translation result. MUNIT~\cite{huang2018munit} is a GAN-based model that uses latent embedding reconstruction losses to disentangle appearance from structure.  VQI2I~\cite{chen2022eccv} uses a vector-quantized GAN to encode images into sequences of tokens representing the structure and appearance of the input images, and uses embedding reconstruction losses to enforce the disentanglement of the structure.  Since these methods are not class-conditional, we trained them on each class separately. We report the average result across the examples of all classes as well as the results of training on the entire datasets. 

Although {\MethodName} does not utilize paired data, we also consider as baseline the state-of-the-art \emph{paired} sketch-to-photo  method CoGS~\cite{ham2022cogs}. 
We note that VQI2I, CoGS and MUNIT allow diverse sampling with an additional appearance image or vector as input, whereas {\MethodName} samples diverse results by varying the random seed.

\begin{figure*}[ht]
    \centering
    \includegraphics[width=0.95\linewidth]{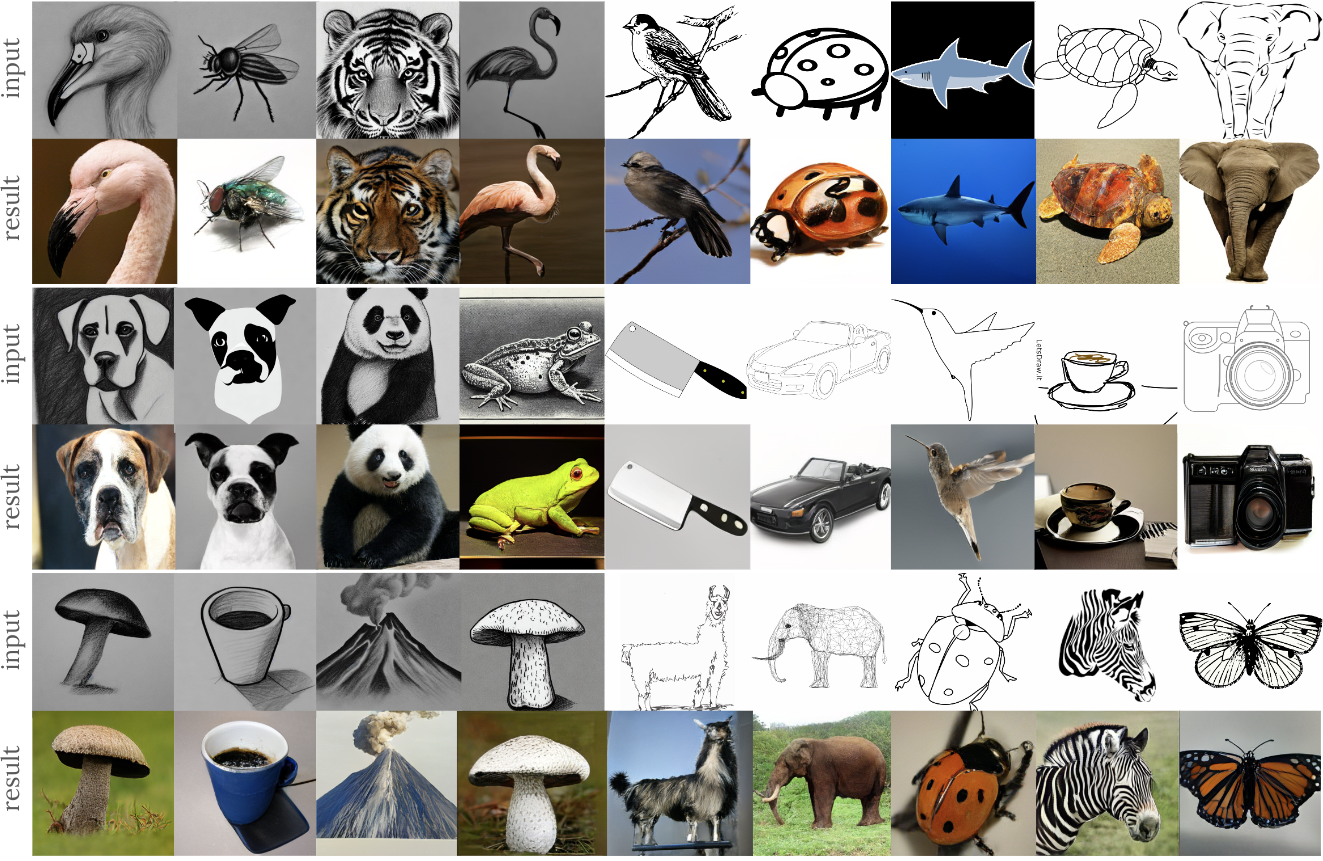}
    \caption{Example  translations by {\MethodName} on the \DatasetName~ dataset. The model takes as input a sketch and a class label.  }
    \label{fig:examples}
\end{figure*}

\paragraph{Datasets}
For qualitative evaluation of \MethodName, we propose \DatasetName, a novel dataset made of 200 openly licensed  sketches. \DatasetName~ contains real sketches drawn with pencil and paper, as well as digital ske`tches. Furthermore, to mimick realistic, highly detailed sketches, we utilized the open source implementation of Stable Diffusion~\cite{rombach2022high} to generate input examples.
All the sketches shown in this manuscript are from \DatasetName.

For quantitative evaluation, we considered two datasets: ImageNet-Sketch~\cite{wang2019learning}, a dataset of 50 real sketches of 1000 classes of ImageNet-2012 \cite{deng2009imagenet} and the    Pseudosketches dataset~\cite{ham2022cogs}, consisting of pairs of ground truth real images and their corresponding automatically extracted edge maps from 125 classes from the ImageNet21K~\cite{ridnik2021imagenet} dataset. We present qualitative results for these datasets in~\cref{sec:sup_main_res}\footnote{Not shown in the main manuscript due to copyright concerns.}.
In our quantiative experiments, we report the results on two versions of the datasets: 1) \emph{full}: using all examples from each of the datasets, and 2) \emph{10-class}:  using the 10 classes that are reported to result in the highest-quality translation results in CoGS~\cite{ham2022cogs}:
``songbird'', ``pizza'', ``volcano'', ``zebra'', ``castle'', ``door'', ``shark'', ``mushroom'', ``cup'', ``lion''. The 10-class subsets of     Pseudosketches and ImageNet-Sketch consist of $1{,}749$ and $508$ examples respectively, whereas the full datasets consist of $113{,}370$ examples and $52{,}888$ examples, respectively.
For the 10-class subsets, we trained unpaired image translation baselines that are not class-conditional on each class separately and reported the aggregated results over all 10 classes for a fair comparison with the class-conditional \MethodName. For the full version of the datasets, we train the baseline methods on all classes without class conditioning. 
Since ImageNet-Sketch does not provide ground truth paired data, it is impossible to train CoGS~\cite{ham2022cogs} on this dataset, therefore we use the model trained on     Pseudosketches for both datasets. 


\vspace{-10pt}\paragraph{Metrics}
Quantitative evaluation of sketch-to-photo translation consists of two aspects: evaluation of realism of the generation results, and evaluation of structure fidelity with respect to the input sketch. To estimate realism, we use the FID score~\cite{heusel2017gans}. To assess generation diversity, we used the LPIPS-based diversity score~\cite{ojha2021few}, which computes the average LPIPS~\cite{zhang2018perceptual} distance between the generated examples. For a fair comparison and due to the limited number of samples in the 10-class subsets, we report the FID and LPIPS results over $10{,}000$ examples generated with different `appearance' inputs with the baseline methods CoGS, MUNIT and VQI2I, and with different seeds for \MethodName. For CUT, we diversify the generated set with augmentations. The FID score is computed with respect to the images from ImageNet~\cite{deng2009imagenet} for the ImageNet-Sketch experiments, and with respect to the ground truth Pseudosketches images for the Pseudosketches experiments.

To provide additional quantitative evaluation of structure preservation quality and realism, we also report the two CLIP-based metrics defined in \cref{sec:fidelity_realism}: image feature distance and prompt similarity score. 
The CLIP feature distance metric is more appropriate for the evaluation of structure preservation quality than the edge-based metrics~\cite{ham2022cogs} since the CLIP features are more invariant to the domain gap as shown in the recent works on image-to-sketch translation~\cite{vinker2022clipasso}. 
We note that these metrics are identical to the CLIP-based rejection sampling in~\cref{sec:method}, and we include the quantitative results without CLIP-based sampling in~\cref{sec:sup_clip_selection}.

\vspace{-10pt}\paragraph{User Preference Studies}
Quantitative evaluation of structure fidelity is challenging due to the distribution shift between the shapes of real objects and abstract sketches and outlines. To complement the quantitative results, we performed user preference studies. 
We asked users the question: \textit{"Given the task of converting the sketch shown on the left into a realistic photo, which result do you prefer?"}. Users were asked to pick one result among the five compared methods (CoGS, MUNIT, VQI2I, CUT and \MethodName) according to their preference. We collected three preference evaluations for each example in the 10-class ImageNet-Sketch and  Pseudosketches datasets. 
Finally, we counted only unanimous votes to guarantee statistical significance. Please see ~\cref{sec:sup_user} for more details.


\section{Results}
\label{sec:quant}

\paragraph{Quantitative Results} In~\cref{tab:quantitative_10class}, we report the quantitative evaluation results on the 10-class subsets of ImageNet-Sketch and Pseudosketch. Additionally, we report the FID and LPIPS diversity scores over the full ImageNet-Sketch and Pseudosketch datasets in~\cref{tab:imagenet_sketch_all}. The results on the 10-class subsets indicate an advantage of {\MethodName} in terms of realism and diversity, with a two-fold decrease in the FID score compared to the baseline MUNIT on both ImageNet-Sketch and Pseudosketch datasets. In our experiments, {\MethodName} outperformed the baselines on the entire ImageNet-Sketch dataset of real sketches, including the fully-supervised CoGS on the Pseudosketches. Notably, general image translation methods, such as MUNIT and CUT, outperform the fully-supervised CoGS in a class-supervised setup. As seen from the FID and LPIPS results, VQI2I struggles to generalize on the relatively small ImageNet-Sketch dataset that contains only 50 examples in each class, mainly due to mode collapse. 

\begin{table*}[t]
    \centering
    \begin{tabular}{l |c| c c | c c | c  }
        \toprule
         & supervision & FID $\downarrow$ & LPIPS $\uparrow$ & CLIP prompt  $\uparrow$ & CLIP feat. $\downarrow$ &
          User preference  $\uparrow$ \\\midrule
           \multicolumn{7}{c}{ImageNet-Sketch} \\ \midrule
        MUNIT & class & $68.65$ & $0.58$ & $52.10$ & $30.27$ &  $10.70\%$ \\
        CUT & class & $77.74$ & $0.68$ & $65.59$ & $28.05$ & $19.78\%$ \\
        VQ-I2I & class & $181.77$ & $0.32$ & $53.76$ & $31.05$& $0\%$  \\
        CoGS & class + pairs & $97.31$ & $0.64$ & $56.62$ & $29.52$ & $8.55\%$ \\
         {\MethodName}(ours) & class & $\bm{33.24}$ & $\bm{0.78}$ & $\bm{67.10}$ & $\bm{26.63}$ & $\bm{59.35\%}$ \\ \midrule
         \multicolumn{7}{c}{Pseudosketches} \\ \midrule
         MUNIT & class & $93.23$ & $0.69$ & $41.91$ & $27.65$ & $23.08\%$ \\
        CUT & class & $112.11$ & $0.42$ & $45.67$ & $27.62$ & $25.0\%$ \\
        VQ-I2I & class & $169.1$ & $0.77$ & $34.5$ & $28.47$ & $0.64\%$ \\
         CoGS & class + pairs & $102.66$ & $0.68$ & $34.79$ & $27.52$ & $14.10\%$ \\
         {\MethodName} (ours) & class & $\bm{56.55}$ & $\bm{0.78}$ & $\textbf{59.48}$ & $\textbf{25.60}$ &  $\bm{35.25\%}$ \\
        \bottomrule
    \end{tabular}
    \caption{Sketch-to-photo translation performance on ImageNet-Sketch (\textbf{top}) and Pseudosketches (\textbf{bottom}) 10-classes subsets.
}
    \label{tab:quantitative_10class}
\end{table*}

\paragraph{Qualitative Results}
\label{sec:qual}
Qualitative comparison shows that the baseline image translation methods, including the supervised CoGS, capture the overall layout and outlines of the input sketch but sometimes fail to produce realistic results. For instance, the GAN-based architectures, namely CUT and MUNIT, produce structurally similar results by practically recoloring the input sketch, which results in a sub-par realism, especially on the more abstract sketches. In our experiments, the VQ-GAN-based VQI2I model failed to learn the correspondences between hand-drawn sketches from ImageNet-Sketch and images from the real photo domain due to a limited number of examples in ImageNet-Sketch, therefore we observe a severe mode collapse on most classes. The fully-supervised CoGS sometimes failed to produce realistic and semantically meaningful results, especially on the hand-drawn sketches. {\MethodName} achieved a good balance between realism and structure fidelity on the majority of sketches from ImageNet-Sketch. However, {\MethodName} struggled to preserve structure on some examples from Pseudosketches due to the extreme complexity of the extracted edge maps.



\begin{figure}[h]
    \centering
 \includegraphics[width=\linewidth]{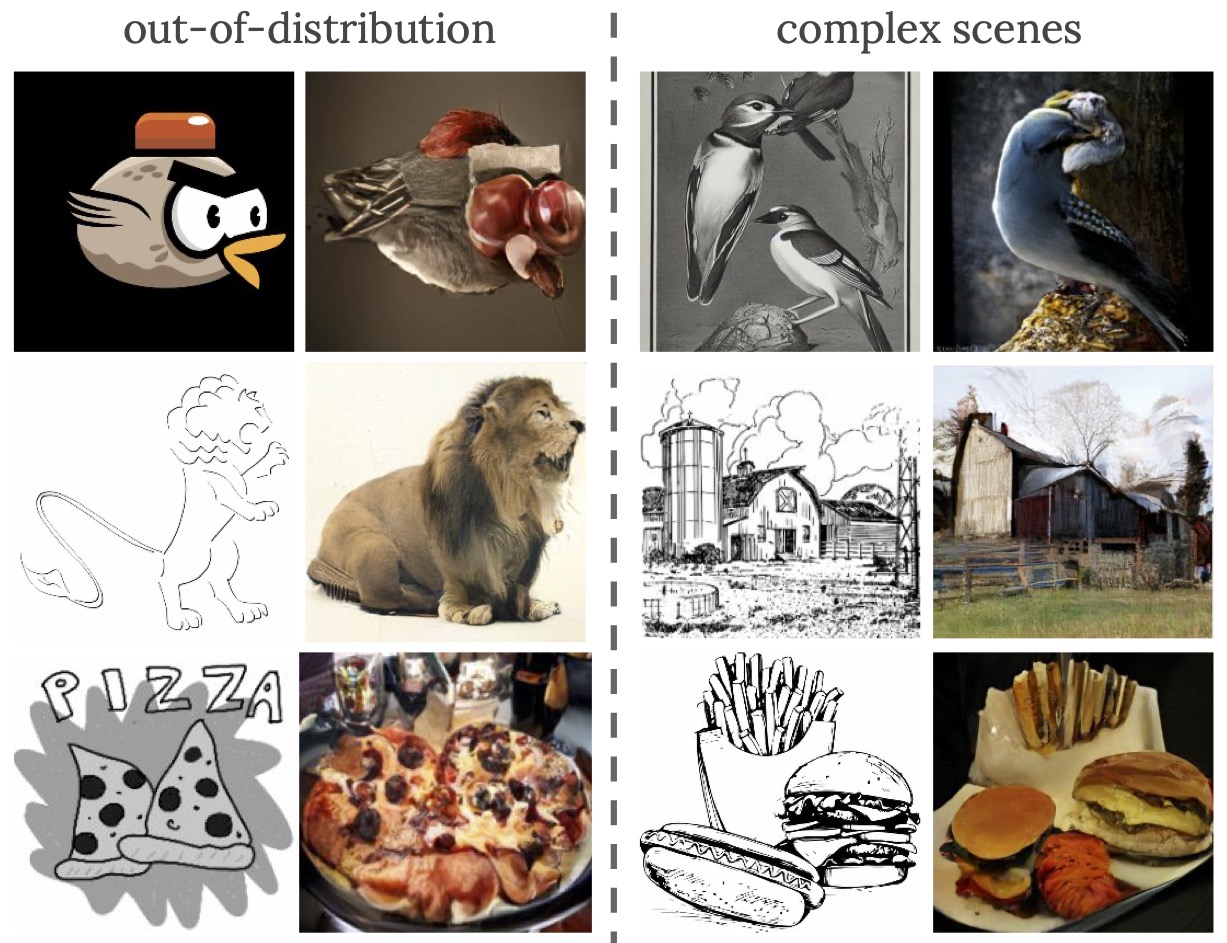}
    \caption{Failure cases of {\MethodName} on the hand-drawn sketch examples: out-of-distribution composition and shapes, complex scenes containing multiple objects. }
    \label{fig:failure}
\end{figure}

\paragraph{Limitations}
The main limitation of \MethodName~ is computational efficiency. To achieve a successful optimization of the structural constraint, {\MethodName~} requires significantly more sampling iterations than the regular MaskGIT. Furthermore, to improve the stability of results it was necessary to apply a multiple trials rejection scheme. 
Two other important limitations for \MethodName~ are the coarse granularity of the attention maps in a transformer, and the flexibility of the prior model, in our case an ImageNet-pretrained MaskGIT.
Figure~\ref{fig:failure} illustrates the common failure cases of our method: out-of-distribution scene composition scarcely or not represented in the training set of MaskGIT, multiple objects forming an unrealistic scene, as well as the complex scenes with multiple foreground and background objects.

\begin{table}[]
    \centering
    \begin{tabular}{l  c c c c  }
    \toprule
    & \multicolumn{2}{c}{ImageNet-Sketch} & \multicolumn{2}{c}{Pseudosketches} \\
    \cmidrule(r){2-3}\cmidrule(l){4-5}
         & FID $\downarrow$ & LPIPS $\uparrow$ & FID $\downarrow$ & LPIPS $\uparrow$ \\\midrule
        MUNIT & $113.45$  & $0.74$ & $121.69$ & $0.71$ \\
        CUT & $161.33$ & $0.74$ & $163.82$ & $0.70$  \\
        VQI2I & $131.70$ & $0.72$ & $135.47$ & $0.71$ \\
        CoGS (sup.) & $85.09$  & $0.72$ & $49.31$ & $0.71$ \\
        \MethodName & $\textbf{23.89}$ & $\textbf{0.77}$ & $\textbf{46.44}$ & $\textbf{0.78}$ \\
        \bottomrule
    \end{tabular}
    \caption{Comparison on the full ImageNet-Sketch (\textbf{left}) and Pseudosketches-validation (\textbf{right}) datasets. }
    \label{tab:imagenet_sketch_all}
\end{table}

\section{Conclusion}
\label{sec:conclusion}
We proposed \MethodName, a sketch-guided image generation method that allows control over the spatial layout of the generation result. {\MethodName} achieves high realism and structure preservation without pairwise supervision, does not require model finetuning and works on sketches of various levels of abstraction. We show that the self-attention maps of the intermediate layers of a masked generative transformer encode important structural information of the input image and are sufficiently domain-invariant, which allows their use in a structure similarity constraint. Our experimental results show that the proposed attention-based sampling approach outperforms state-of-the-art sketch-to-photo and general image translation methods in terms of both realism and structure fidelity. 

\paragraph{Acknowledgements}
We thank Tali Dekel, Huiwen Chang, Lu Jiang, and David Salesin for their insightful advice and guidance. This work was done during an internship at Google Research.
\newpage



{\small
\bibliographystyle{ieee_fullname}
\bibliography{refs}
}

\appendix
\newpage
\clearpage
\section{Method implementation details}
\label{sup:method}
{\MethodName} is implemented in Jax~\cite{jax2018github} / Flax~\cite{flax2020github} similarly to the official implementation of MaskGIT. We will release the implementation of {\MethodName} upon acceptance. We used an ImageNet-pretrained $256{\times}256$ VQGAN encoder-decoder and a 24-layer BERT transformer in all experiments.\footnote{The VQGAN and transformer model checkpoints used in our experiments are found in \url{https://github.com/google-research/maskgit}.} In all experiments, we used the following parameters:
\begin{itemize}
    \item layers $1, 3, 16, 18, 20, 21, 22$ for the structure distance objective in ~\cref{eq:structure_dist} 
    \item Gumbel temperature $0$ for ImageNet-Sketch and $0.001$ for Pseudosketches experiments.
    \item 4 sampling trials for ImageNet-Sketch and 3 sampling trials for the Pseudosketches.
    \item $1000$ iterations for ImageNet-Sketch and $ 500$ iterations for Pseudosketches.
    \item Classifier-free guidance scales of $(0., 0.1, 0.25, 0.5)$ for ImageNet-Sketch and $(0., 0.05, 0.1)$ for Pseudosketches, varied for each iteration trial accordingly.
    \item $\lambda_s$ is set to $0.9$ for ImageNet-Sketch and to $0.95$ for Pseudosketches.
    \item Starting mask rate is set to $0.95$ for both datasets, and the end mask rate is $0.25$ for ImageNet-Sketch and $0.33$ for Pseudosketches.
    \item Token-Critic parameters: We used the Token-Critic refinement ratio $r_{tc}=0.5$ and $r_{tc}=0.6$ for ImageNet-Sketch and Pseudosketches experiments, respectively, and set the number of refinement steps to $N_{tc}=32$ (explained in ~\cref{sec:token_critic}).
\end{itemize}

\begin{table}[]
    \centering
    \begin{tabular}{l|c c c}
    \toprule
         & $3/3$ & $2/3$ & Overall  \\ \midrule
          \multicolumn{4}{c}{ImageNet-Sketch 10-class} \\ \midrule
        MUNIT & $10.70\%$ & $13.44\%$ & $13.80\%$ \\
        CUT & $19.78\%$ & $24.90\%$ & $21.42\%$ \\
        VQ-I2I & $0\%$ & $0.\%$ & $0.33\%$ \\
        CoGS & $8.55\%$ & $16.60\%$ & $15.24\%$ \\
        \MethodName~(ours) & $59.35\%$ & $40.71\%$ & $44.22\%$ \\
        \midrule
          \multicolumn{4}{c}{Pesudosketches 10-class} \\ \midrule 
        MUNIT & $23.08\%$ & $22.88\%$ & $22.23\%$ \\
        CUT & $25.00\%$ & $24.94\%$ & $23.57\%$ \\
        VQ-I2I & $0.64\%$ & $2.75\%$ & $4.75\%$ \\
        CoGS & $14.10\%$ & $16.02\%$ & $16.76\%$ \\
        \MethodName~(ours) & $35.25\%$ & $31.35\%$ & $27.96\%$ \\
          \bottomrule
    \end{tabular}
    \caption{User preference study: ratios of unanimous votes ($3/3$), exactly two out of three votes ($2/3$) as well as the overall preference on the 10-class subsets of ImageNet-Sketch and Pseudosketches datasets. }
    \label{tab:human_eval}
\end{table}

\begin{table}[]
    \centering
    \begin{tabular}{l|c c c}
    \toprule
         & $3/3$ & $2/3$ & Overall  \\ \midrule
          \multicolumn{4}{c}{ImageNet-Sketch 10-class} \\ \midrule
        MUNIT & $20$ & $34$ & $210$ \\
        CUT & $37$ & $63$ & $326$ \\
        VQ-I2I & $0$ & $0$ & $5$ \\
        CoGS & $16$ & $42$ & $232$ \\
        \MethodName~(ours) & $111$ & $103$ & $673$ \\
        No selection & $3$ & $11$ & $76$ \\
        \midrule
          \multicolumn{4}{c}{Pesudosketches 10-class} \\ \midrule 
        MUNIT & $36$ & $100$ & $528$ \\
        CUT & $39$ & $109$ & $560$ \\
        VQ-I2I & $1$ & $12$ & $113$ \\
        CoGS & $22$ & $70$ & $398$ \\
        \MethodName~(ours) & $55$ & $137$ & $664$ \\
         No selection & $3$ & $9$ & $112$ \\
          \bottomrule
    \end{tabular}
    \caption{User preference study: number of the unanimous votes ($3/3$), exactly two out of three votes ($2/3$) as well as the overall number of votes on the 10-class subsets of ImageNet-Sketch and Pseudosketches datasets. The participants were asked to select the ``No selection'' option on the examples on which all methods performed comparatively poorly or the sketch content was unclear. We excluded the ``No selection'' examples from the statistics in ~\cref{tab:human_eval} and ~\cref{tab:quantitative_10class}. }
    \label{tab:human_eval_raw}
\end{table}

\section{Structure-guided sampling}
\label{sec:sup_attn_res}
\begin{figure*}[p]
    \centering
    \includegraphics[width=\textwidth]{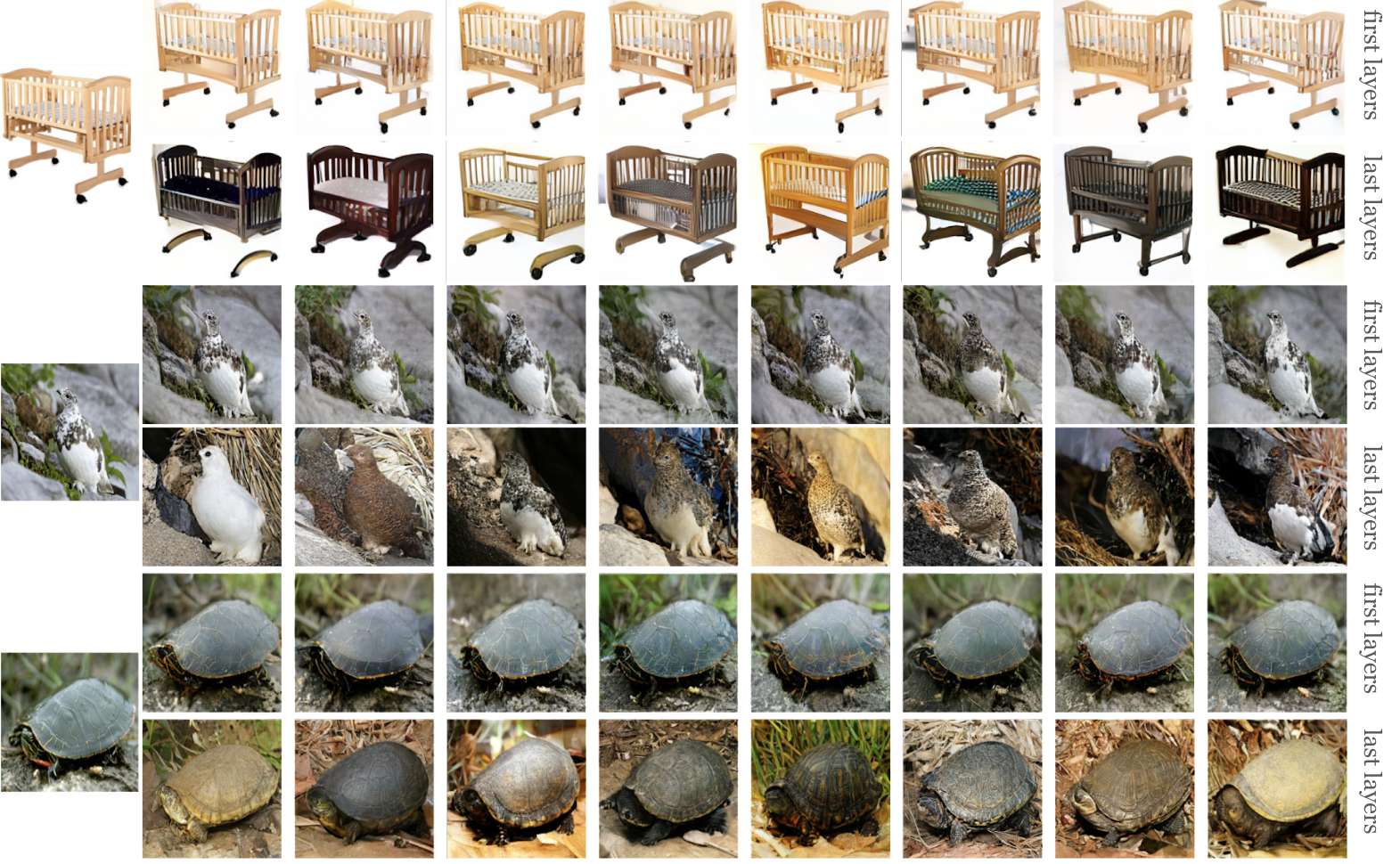}
    \caption{Structure-guided sampling examples using layers $\mathcal{L}=\{1,2,3\}$ (top of each row) and layers $\mathcal{L}=\{16, 18, 20\}$ (bottom of each row).}
    \label{fig:sup_attn_sampling}
\end{figure*}

Please see ~\cref{fig:sup_attn_sampling} for more examples of the structure-guided sampling across the first and last layers of MaskGIT (extending \cref{fig:sampling_example_towels}).

\section{Results on ImageNet-Sketch and Pseudosketches}
\label{sec:sup_main_res}
 Unfortunately, we cannot include the illustration on ImageNet-Sketch and Pseudosketches in the main manuscript due to copyright concerns.

\subsection{Ablation of CLIP-based rejection}
\label{sec:sup_clip_selection}

\begin{table}[]
    \centering
    \begin{tabular}{l|c c | c c }
    \toprule
         & FID $\downarrow$ & LPIPS $\uparrow$ & CLIP pt. $\uparrow$ & CLIP ft. $\downarrow$ \\ \midrule
          \multicolumn{5}{c}{ImageNet-Sketch 10-class} \\ \midrule
         No sel. & $34.23$ & $0.77$ & $\bm{71.87}$  & $27.17$ \\
        4-trial sel. &  $\bm{33.24}$ & $\bm{0.78}$ & $67.10$ & $\bm{26.63}$ \\\midrule
         \multicolumn{5}{c}{Pseudosketches 10-class} \\ \midrule
          No sel. & $60.44$ & $0.78$ & $56.31$ & $26.85$ \\
        3-trial sel. & $\bm{56.55}$ & $\bm{0.78}$ & $\textbf{59.48}$ & $\textbf{25.60}$  \\ \bottomrule
    \end{tabular}
    \caption{CLIP-based rejection sampling ablation study.
    \emph{No-sel.} indicates no rejection sampling was used. \emph{4-trial sel. and 3-trial sel.} indicates  selecting one sample out of 4 and 3 trials, respectively.
    \emph{CLIP pt.} is the CLIP prompt similarity between the translation result and the prompt \textit{"Photo of a \textbf{c}"}, where \textbf{c} is the input class name
    \emph{CLIP ft.} is the CLIP feature distance between the input sketch and the corresponding translation. \vspace{-10pt}}
    \label{tab:clip_ablation}
\end{table}

\section{Token-Critic refinement}
\label{sec:token_critic}
In our experiments, we used the ImageNet-trained Token-Critic~\cite{lezama2022token} refinement to further improve realism of the translation results. In Token-Critic refinement, the tokens of a sampled image are passed to a critic transformer model that outputs a conditional likelihood score for each token. The score is high for tokens that are likely under the data distribution and low otherwise. We refine a sampled image by using the Token-Critic scores as the confidence scores in \cref{algo:sampling}, and setting $\lambda_s=0$ (no structure guidance). The refinement process uses a mask rate of $r_{tc}$.
We used $r_{tc}=0.5$ and $r_{tc}=0.6$ for ImageNet-Sketch and Pseudosketches experiments, respectively, and set the number of refinement steps to $N_{tc}=32$, and in both experiments, the mask ratio varies across iterations according to the cosine schedule.

\section{User preference study}
\label{sec:sup_user}
For all the validation images in the ImageNet 10-classes and Pseudosketches 10-classes datasets, we asked the participants to pick one option that best answers the question: \textit{``Given the task of converting the sketch shown on the left into a realistic photo, which result do you prefer?''}. For each example, we got the answers from three participants, and we report the unanimous voting results ($3/3$) in ~\cref{tab:quantitative_10class}. We report the ratios of choices of the user preference study in ~\cref{tab:human_eval}: statistics for the unanimous votes ($3/3$), exactly two out of three votes ($2/3$) as well as the overall preference. We also report the total number of choices in~\cref{tab:human_eval_raw}.


\section{CLIP-based metrics}
\label{sec:sup_clip_metrics}

\paragraph{Structure distance}
To estimate structure similarity between the input sketch $\bm{x}$ and the translation result $\bm{y}$, we compute $L_1$-distance between the ResNet101-based CLIP image encoder intermediate layer features: $\operatorname{CLIP}^{s}(\bm{x}, \bm{y}) = ||\operatorname{CLIP}_{l}(\bm{x}) - \operatorname{CLIP}_{l}(\bm{y})||_1$, where $l$ is the ResNet-101 layer block index. In our experiments, we use the last layer block ($l=4$). 

\paragraph{Prompt similarity}
To asses realism and semantic accuracy of the translation result, we use CLIP zero-shot classification to estimate the relative similarity between the translated image and the prompt ``\textit{Photo of a \textbf{c}}", where $c$ is the ground truth class label index corresponding to the input sketch. Therefore, given an input sketch $\bm{x}$ of class $c$, the prompt similarity is computed as:
\[
\operatorname{CLIP}^{r}(c, \bm{y}) = \operatorname{softmax}\{CLIP(\bm{y})^T CLIP(\bm{p})\}[c]
\]
where $\bm{p} = [\text{``Photo of a \textbf{m}"} \text{ } \forall \bm{m} \in \Omega]$, $\Omega$ is the set of class labels in the dataset.

\section{Comparison with PITI}
\label{sec:piti}
In this section, we provide the quantitative and qualitative comparison with the concurrent \emph{supervised} image-to-image translation method PITI~\cite{wang2022pretraining}. For a fair comparison, we compared the generation results on the four classes from the intersection of classes of the MS COCO~\cite{lin2014microsoft} dataset that was used to train PITI and ImageNet-Sketch 10 classes we used to compare with the other baseline methods. Since PITI is sensitive to the modality of the input (e.g., it produces subpar results on inverted sketches), we used PITI
's edge extraction pipeline on the input sketches before translating with PITI. The CLIP-based evaluation results on ~\cref{tab:piti} show that PITI results are slightly better in terms of structure fidelity, however they are generally less realistic than \MethodName~ translation results. An important disadvantage of PITI is its sensitivity to the domain shift: the edge extraction method HED~\cite{xie2015holistically} that was used to train PITI removes some edges in the given sketch, which results in errors in structure and even misclassification of the input sketch (e.g. PITI typically confuses the round pizza shape with other round objects, such as watch or bowl). 

\begin{table}[]
    \centering
    \begin{tabular}{c| c c}
    \toprule
         & CLIP ft. & CLIP pt. \\
         \midrule
        PITI & $\bm{25.0}$ & $59.1$ \\
        MaskSketch (ours) & $27.3$ & $\bm{68.2}$ \\
        \bottomrule
    \end{tabular}
    \caption{CLIP-based evaluation (\cref{sec:fidelity_realism})  on 4 classes from the intersection of classes in MS-COCO~\cite{lin2014microsoft} and ImageNet-Sketch 10-class datasets: zebra, pizza, songbird, door. }
    \label{tab:piti}
\end{table}

\end{document}